\documentclass[10pt,twocolumn,letterpaper]{article}

\usepackage{cvpr}              % To produce the CAMERA-READY version
\usepackage{multirow}
\usepackage{wrapfig} 
\usepackage[accsupp]{axessibility}
\usepackage[dvipsnames]{xcolor}

\definecolor{cvprblue}{rgb}{0.21,0.49,0.74}
\usepackage[pagebackref,breaklinks,colorlinks,citecolor=cvprblue]{hyperref}

\def\paperID{5619} % *** Enter the Paper ID here
\def\confName{CVPR}
\def\confYear{2024}

\title{UniGarmentManip: A Unified Framework for Category-Level Garment Manipulation via Dense Visual Correspondence}

\author{
Ruihai Wu\footnotemark[1] $^{1,4}$ \quad 
Haoran Lu\footnotemark[1] $^{2,1}$ \quad
Yiyan Wang $^{3,1}$ \quad
Yubo Wang $^{2,1}$  \quad 
Hao Dong\footnotemark[2] $^{1,4}$ \quad  \\
$^1$CFCS, School of CS, PKU \quad 
$^2$School of EECS, PKU \quad 
$^3$School of CS\&T, BIT
\\$^4$National Key Laboratory for Multimedia Information Processing, School of CS, PKU
}

\begin{document}
\maketitle
\renewcommand{\thefootnote}{\fnsymbol{footnote}}
\footnotetext[1]{Equal contribution.}
\footnotetext[2]{Corresponding author.}

\begin{abstract}
Garment manipulation (\emph{e.g.}, unfolding, folding and hanging clothes) is essential for future robots to accomplish home-assistant tasks, while highly challenging due to the diversity of garment configurations, geometries and deformations.
Although able to manipulate similar shaped garments in a certain task,
previous works mostly have to design different policies for different tasks, could not generalize to garments with diverse geometries, and often rely heavily on human-annotated data.
In this paper, we leverage the property that,
garments in a certain category have similar structures,
and then learn the topological dense (point-level) visual correspondence among garments in the category level with different deformations in the self-supervised manner. 
The topological correspondence can be easily adapted to the functional correspondence to guide the manipulation policies for various downstream tasks,
within only one or few-shot demonstrations.  
Experiments over garments in 3 different categories on 3 representative tasks in diverse scenarios, 
using one or two arms, 
taking one or more steps,  
inputting flat or messy garments,
demonstrate the effectiveness of our proposed method.
Project page: \href{https://warshallrho.github.io/unigarmentmanip}{https://warshallrho.github.io/unigarmentmanip}.

\end{abstract}
\vspace{-5mm}
\section{Introduction}

Next-generation robots should have the abilities to manipulate a large variety of objects in our daily life,
including rigid objects, articulated objects~\cite{du2023learning, geng2023gapartnet, geng2023partmanip} and deformable objects~\cite{wu2023learning}.
Compared with rigid and articulated objects,
deformable objects are much more difficult to manipulate,
for the highly large and nearly infinite state and action spaces, and complex kinematic and dynamics.
Garments, such as shirts and trousers, are essential types of deformable objects,
due to the potentially wide-range applications for both industrial and domestic scenarios.
Manipulating garments, such as unfolding, folding and dressing up, has garnered significant interest in robotics and computer vision fields.

There have been a long range of studies on manipulating
relatively simple shaped deformable objects,
such as square-shaped cloths~\cite{wu2023learning, wu2020learning, lin2021VCD}, ropes and cables~\cite{wu2023learning, wu2020learning, seita2021learning}, and bags~\cite{bahety2022bag, chen2023autobag}.
Nevertheless,
manipulating garments presents a substantial challenge,
as it necessities the comprehensive understanding of more diverse geometries (garments in a certain category have different shapes, let alone in different categories), more complex states (various geometries with diverse self-deformations),
and more difficult goals (\emph{e.g.}, garments require multiple fine-grained actions fold step by step).
Many existing studies on garment manipulation rely on large-scale annotated data~\cite{canberk2022clothfunnels, avigal2022speedfolding}, which is labor-intensive and time-consuming, hindering the scalability in the scenarios of real-world applications.
Besides, many works design quite different methods to tackle different specific tasks~\cite{Wang2023One, canberk2022clothfunnels, xue2023unifolding, avigal2022speedfolding},
making it difficult to efficiently share and reuse information among different tasks.

\begin{figure*}[t]
  \centering
  \vspace{-7mm}
  \includegraphics[width=1\textwidth]{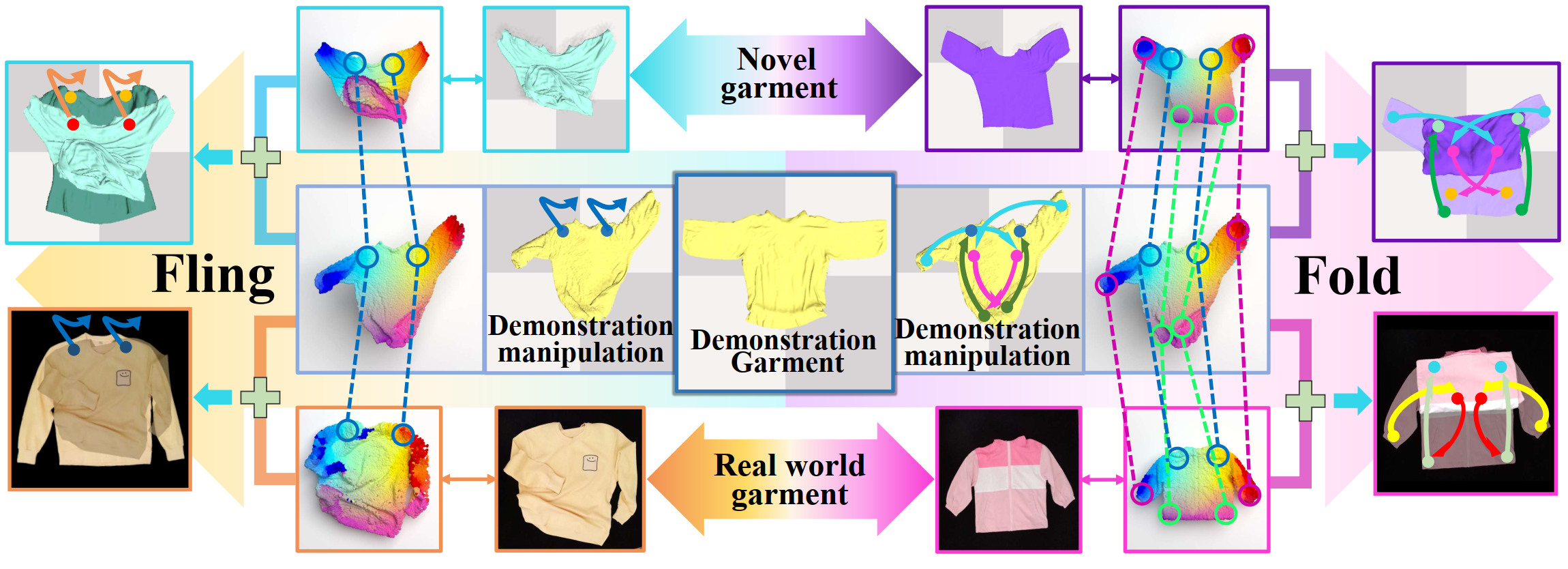}
  \vspace{-7mm}
   \caption{Given a demonstration garment \textbf{(Middle)} and the demonstration actions to fulfill a task \textbf{(Middle-Left/-Right)}, for a novel object, we find the manipulation points using the proposed \textbf{Dense Visual Correspondence for Garment Manipulation} and execute the corresponding action to fulfill the task \textbf{(Left/Right)}. Color similarity denotes in the correspondence space.}
   \vspace{-2mm}
   \label{fig:teaser}
\end{figure*}

Different from other object types, garments possess a property that,
in a certain category,
while different garments may have different geometries,
they usually share the same structure.
For example,
tops (such as T-shirts, jackets and jumpers),
are composed of certain components (a body with two sleeves and a collar),
and the topological structures of the components are usually the same,
even though the length, width and geometries of a certain component in different garments may be quite different.
Thanks to such similarity in structure shared among garments in the category level,
it is easy for humans to fulfill a task on unseen novel garments using the experience of manipulating only one or a few garments in the same category.
Therefore,
we empower robots with the above one/few-shot generalization ability humans have in diverse tasks, by
leveraging such structural similarity among garments.

Among multiple ways to describe and represent garments (\emph{e.g.}, poses~\cite{chi2021garmentnets, xue2023garmenttracking}, lines~\cite{zhu2020deep, gabas2017physical} and keypoints~\cite{zhou2023clothesnet}),
\textbf{skeleton}~\cite{shi2021skeleton}, \emph{i.e.}, a graph of keypoints covering significant points on garment edges and joints to represent the topology of 3D objects, is suitable for describing the above-mentioned structures shared among garments.
The skeleton points are sparse, distinct and ordered, and thus (1) exist on each garment and (2) can easily distinguish with other skeleton points, making them easy to learn.
Therefore,
we use skeleton points to build structural correspondence among garments.
Moreover, as different-extent self-deformations make the garments to be quite complex,
while previous works only studied skeleton points on rigid~\cite{shi2021skeleton}, articulated~\cite{xu2017lie} or fixed-posed deformable objects in the canonical view~\cite{zhou2023clothesnet},
we further extend skeleton points to garments at any deformation states,
making a step to more realistic scenarios for garment manipulation.

While skeletons build topological correspondence between different garments in the skeleton keypoint level,
the state and action spaces of garments are exceptionally large and each point on the garment could be the manipulation point,
making the sparse skeleton points unable to fully represent garments for manipulation.
To represent objects with large state and action spaces,
dense (\emph{i.e.}, point-level or pixel-level) object representations,
including dense object descriptors~\cite{florencemanuelli2018dense} and dense visual actionable affordance~\cite{mo2021where2act},
which indicate the actionable information on each point of the object,
have demonstrated its superiority on rigid~\cite{florencemanuelli2018dense}, articulated~\cite{wu2022vatmart}, and simple-shaped deformable object manipulation~\cite{wu2023learning}.
We further extend dense representations to garments, with the awareness of correspondences, using the proposed skeleton points,
and thus achieve fine-grained manipulation.

With dense visual correspondence aware of garment structures,
one demonstration can roughly guide manipulating a novel garment by indicating  corresponding action points and policies.
Furthermore,
as manipulation for specific tasks rely on not only garment structures but also task-specific knowledge,
we further transform the representation from task-agnostic structural to task-specific functional for more accurate manipulation in various downstream tasks,
using few-shot demonstrations to achieve this adaptation.

To demonstrate the performance of our proposed representations,
we conduct experiments on 3 different kinds of garments over 3 representative tasks.
The experimental results showcase the superiority of our proposed framework in manipulating diverse novel garments in multiple tasks using the proposed dense visual correspondence and one or few-shot demonstrations.

In summary, our contributions include:

\begin{itemize}
\item We propose to learn category-level dense visual correspondence to reflect the topological and functional correspondence across garments in different styles or deformations,
which is an unified representation that  facilitates manipulating diverse unseen garments in multiple tasks with one or few-shot demonstrations.
\item 
We propose a novel learning framework with novel designs to efficiently learn the proposed dense visual representation for garments.
\item
Experiments over diverse representative tasks demonstrate the effectiveness of our proposed dense correspondence and the learning framework.
\end{itemize}

\vspace{-3mm}
\section{Related Work}

\subsection{Dense Representations for Manipulation}

Dense object descriptors~\cite{florencemanuelli2018dense} that learn point- or pixel-level object representations are proposed by and for robotic manipulation. Many works extend such descriptors to propose grasp pose~\cite{yen2022nerf, simeonovdu2021ndf},  manipulate ropes~\cite{sundaresan2020learning}, smooth fabrics~\cite{ganapathi2021learning}.
Another series of works learn point-level affordance for articulated ~\cite{li2024unidoormanip, ling2024articulated} and deformable object~\cite{wu2023learning}, language-guided~\cite{xu2024naturalvlm}, mobile~\cite{li2024mobileafford} and bimanual~\cite{zhao2022dualafford} manipulation, with exploration for interaction~\cite{wang2021adaafford, ning2023where2explore},
facilitating point-level contact point selection for diverse downstream tasks.
Our dense correspondence extends dense object representations in  garment manipulation.

\subsection{Visual Correspondence Learning}

Learning visual correspondence aims to reflect the shared information (\emph{e.g.}, geometric, topological and functional information) between different objects, which facilitates generalization in diverse tasks,
including functional perception~\cite{lai2021functional}, 
pose estimation~\cite{haugaard2022surfemb},
grasping~\cite{yen2022nerf, patten2020dgcm, xue2022useek, ding2024preafford} and 
fabric manipulation~\cite{ganapathi2021learning}.
For garments,
although different garments have quite different geometries and deformations in different states,
they share similar structural and topological information in the category level, which can help in manipulating novel garments with the demonstration of a garment with the similar structure.
So
we propose to learn correspondence between garments to
facilitate novel garment manipulation in diverse downstream tasks.

\subsection{Cloth and Garment Manipulation}

Manipulating a square-shaped cloth is relatively well-studied, with previous works leveraging flow and dynamics~\cite{weng2021fabricflownet, lin2021VCD}, tactile feedback~\cite{tirumala2022reskin, pmlr-v205-sunil23a}, dense representations~\cite{wu2023learning, seita2021learning} and reinforcement learning~\cite{jangir2020dynamic, matas2018sim} to tackle different tasks.
Garment manipulation is more challenging, for the diversity of garment types and shapes,
requiring the method to handle diverse objects and states.
While previous works mainly learn the policy for a certain task, such as folding~\cite{avigal2022speedfolding, canberk2022clothfunnels,xue2023unifolding}, unfolding~\cite{li2015regrasping}, grasping~\cite{chen2023learning, zhang2020learning} and dressing-up~\cite{Wang2023One, doi:10.1126/scirobotics.abm6010}, on similar shaped garments,
we focus on learning garment representations that can generalize to diverse objects in a category and facilitate many tasks.

\vspace{-1.9mm}
\section{Problem Formulation}  
\label{sec:pro}

 Given an $N$-point ($N=10,000$) 3D partial point cloud observation of a garment $O \in \mathbb{R}^{N \times 3}$,
garment manipulation aims to manipulate the garment by a sequence of $n$ actions to complete different tasks.
As explained in~\cite{seita2021learning, seita2020deep, ganapathi2021learning},
each action $a_i$ includes
grasping at a pick point $p_{pick_i}$, 
pulling to a place point $p_{place_i}$ without changing the orientation of the end-effector.
Additionally, for dual-arm manipulation,
each action $a_i$ includes a pair of pick points $p_{pick_i}^1$, $p_{pick_i}^2$ and corresponding place points $p_{place_i}^1$, $p_{place_i}^2$.

Given two garments $O_1$ and $O_2$, dense correspondence evaluates the correspondence (normalized to $[-1, 1]$) in topology or function between each point pair $(p_1, p_2)$, with $p_1$ from $O_1$ and $p_2$ from $O_2$.

Given a task $T$, a demonstration includes the observation $\hat{O}$ of a demonstration garment, and its corresponding demonstration action (including single and dual-arm actions) sequence $(\hat{p}_{pick_1},\ \hat{p}_{place_1},\ ...,\ \hat{p}_{pick_n},\ \hat{p}_{place_n})$ that can fulfill $T$.
Given the observation $O$ of a new garment to fulfill $T$,
manipulation using dense correspondence first finds the points $(p_{pick_1},\ p_{place_1},\ ...,\ p_{pick_n},\ p_{place_n})$ on $O$,
where $p_{pick_i}$ and $p_{place_i}$ have the best correspondence score to $\hat{p}_{pick_i}$ and $\hat{p}_{place_i}$ among all points on $O$, and then executes the corresponding actions to fulfill the task on $O$.

\begin{figure*}[ht]
  \centering
   \includegraphics[width=1\linewidth]{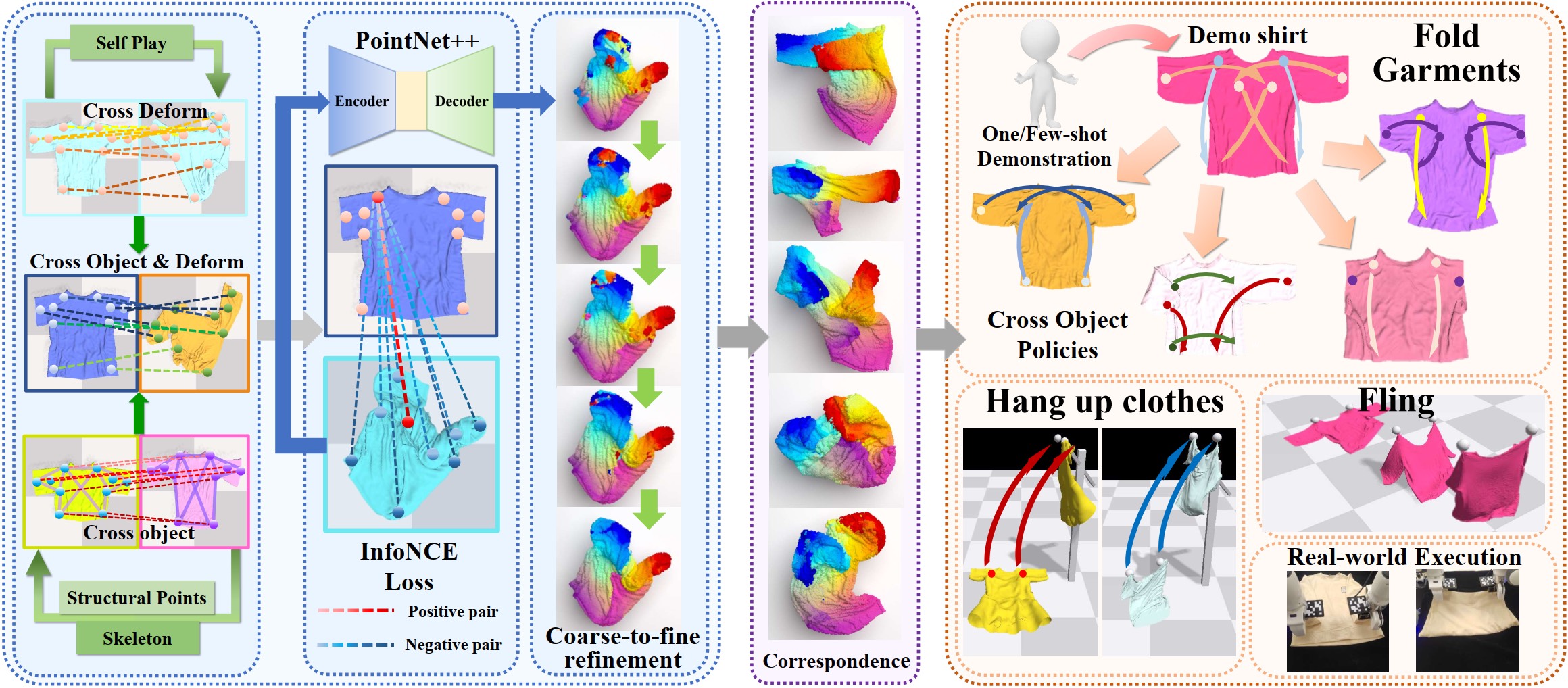}
   \caption{\textbf{Our Proposed Learning Framework for Dense Visual Correspondence.} \textbf{(Left)} We extract the cross-deform correspondence and cross-object correspondence point pairs respectively using self-play and skeletons, and train the per-point correspondence scores in the contrastive manner, with the Coarse-to-fine module refines the quality. \textbf{(Middle)} Learned correspondence demonstrates point-level similarity across different garments in different deformations. \textbf{(Right)} The learned point-level correspondence can facilitates multiple diverse downstream tasks using one or few-shot demonstrations.}
   \label{fig:framework}
\end{figure*}

\section{Method}

\subsection{Overview}

Our framework first learns topological dense visual correspondence aware of different garment deformations and shapes respectively using self-play and skeleton points
 (Section~\ref{sec:method-corr}),
with further coarse-to-fine refinement (Section~\ref{sec:method-c2f}).
After the few-shot adaptation for different downstream tasks, the learned correspondence turns from \textit{topological} to \textit{functional}
(Section~\ref{sec:method-functional}),
and thus could facilitate manipulating unseen novel garments on various tasks using one or few-shot demonstrations (Section~\ref{sec:method-policy}). 
Section~\ref{sec:network}
describes network architectures and the training strategy.
   
\subsection{Self-supervised \textit{Topological} Dense Visual Correspondence Learning}
\label{sec:method-corr}

The diversity of garments in different states mainly comes from two perspectives:
self-deformations, and styles of objects in the same category.
To empower the Dense Visual Correspondence with the alignment ability for different garments in different states,
we decouple the learning process into two parts,
respectively learning
cross-deformation correspondence and cross-object correspondence.

\vspace{-3mm}
\subsubsection{Cross-Deformation Correspondence}
\label{sec:method-cd}

Many tasks, such as unfolding and hanging, 
require manipulating the garment at any random states (\emph{e.g.}, after a random drop).
As demonstrated in~\cite{ganapathi2021learning}, while garments have complex states and infinite deformations,
the manipulation policies (manipulation points) are usually invariant to deformations.
To empower the model with the ability to handle garments in different deformations, 
we introduce learning correspondence across deformations of the same garment.  
 
Given two partial observations $O$ and $O^\prime$ of the same garment in different deformations generated by self-play,
and a visible point $p$ on $O$, we can easily get its corresponding position point $p^\prime$ in $O^\prime$ using point tracing in simulation.
If $p^\prime$ is visible,
the representations $f_{p}$ and $f_{p^\prime} \in \mathbb{R}^{512}$ of $p$ and $p^\prime$ extracted by the backbone network $\mathbf{F}$,
should be the same,
as the representations are agnostic to self-deformations.
We normalize point representations to be unit vectors,
and thus the similarity between $f_{p}$ and $f_{p^\prime}$ can be computed by the dot product of $f_{p}$ and $f_{p^\prime}$, \emph{i.e.}, $f_{p} \cdot f_{p^\prime}$.
For $p$ on $O$,
we use $p^\prime$ on $O^\prime$ as the positive point,
and sample $m$ negative points ($m=150$): $p^\prime_1,\ p^\prime_2,\ ...,\ p^\prime_m$. 
We pull close $f_p$ and $f_{p^\prime}$, while push away $f_p$ and other point representations.
Following InfoNCE~\cite{lai2019contrastive},
a widely-used loss function in one-positive-multi-negative-pair contrastive representation learning,
we identify the positive $p^\prime$ amongst $m$ negative samples:

\begin{equation}
\label{eq:l_cd}
   \mathcal{L}_{CD} = -log(\frac{exp(f_p\cdot f_{p^\prime}/\tau)}{\sum_{i=1}^{m}exp(f_p\cdot f_{p^\prime_i}/\tau)})
\end{equation}
, where $\tau$ denotes the balancing coefficient in InfoNCE.

\subsubsection{Cross-Object Correspondence}
\label{sec:method-co}

In a certain category, while garments highly vary in original shapes, such as sizes, length-width ratios, sleeve lengths and styles,
they share the same topological structure.
The awareness of such structures
will make it easy to manipulate unseen novel garments with demonstrations.

To leverage the shared structural information and generalize to novel shapes,
we propose to use skeleton,
\emph{i.e.}, a graph of keypoints that represents topology of the 3D object,
as the shared bridge for different garments with similar structures.
The reasons for using skeleton include:
\begin{itemize}
    \item 
Skeleton points are \textbf{distinct} and \textbf{sparse}, thus easy to learn and generalize, compared to complicated representations;
    \item 
Skeleton points are \textbf{distinct} and \textbf{ordered}, making it easy to build topological correspondence between two objects by aligning each specific skeleton point on them.
\end{itemize}

To learn garment skeletons in the category level,
we employ the designs of Skeleton Merger~\cite{shi2021skeleton},
which can generate skeletons for rigid objects,
or canonically-posed (\emph{e.g.}, flat) deformable objects.
So we generate skeletons for flat garments.
Specifically, we generate $s\ (s\le 50)$ ordered points on the point cloud observation $O$ as skeleton points, with $s\times(s-1)/2$ activation scores $a_{i,j}$ ($1\le i,j \le s$) indicating whether each edge between 2 skeleton points exists on the object.
We sample points on each edge, with trainable offset for each sample, merging them into the reconstructed object $\bar{O}$.
The training signal is whether $\bar{O}$ covers $O$ (trains skeleton point positions) and whether sampled points on each edge exists in $O$ (trains $a_{i,j}$),
As a result, skeleton points will exist on significant positions on garments (\emph{e.g.}, boundaries, corners and intersections of parts) and meaningful edges will retain, as shown in Figure~\ref{fig:framework} (the Left-Down part).
More implementation details can be seen in the original paper of Skeleton Merger~\cite{shi2021skeleton}.

As skeleton points are \textbf{ordered},
given observation $O$ of a flat garment with one of its skeleton point $p$,
we can get the corresponding skeleton point $\Tilde{p}$ on the observation $\Tilde{O}$ of another flat one, by applying the skeleton network on $\Tilde{O}$ and get the skeleton point in the same order of $p$ in $O$.
Then, the topological correspondence between \textbf{flat garments} have been built in the \textbf{skeleton-point level}.
As the features extracted by neural networks are continuous when point positions continuously change,
and skeleton points cover the whole garment,
the feature of any point can be reflected by its nearby skeleton points (like interpolation) with topological information.
Therefore,
the representation of each point on the garment will reflect its topology,
and  \textbf{dense correspondence} between flat garments has been naturally built.

\subsubsection{Integration of Cross-Deformation and Cross-Object Correspondence}

Since we have designed dense correspondence between \textbf{the same garment in different deformations},
and dense correspondence between \textbf{different flat garments},
the next step is to aggregate them into one dense representation system on diverse garments in any deformation states.

We first project skeletons of garments in their flat states to any deformation states using point tracing in simulation.
Thus,
given the observation $O$ in random deformation with one of its skeleton point $p$,
we can get the corresponding skeleton point $\Tilde{p}$ on the observation $\Tilde{O}$ of another garment in random deformation.
If $\Tilde{p}$ is visible on $\Tilde{O}$,
$f_{p}$ and $f_{\Tilde{p}}$ should be the same.
For $p$ on $O$,
we use $\Tilde{p}$ on $\Tilde{O}$ as the positive point,
and sample $m$ negative points ($m=150$): $\Tilde{p}_1^\prime,\ \Tilde{p}_2^\prime,\ ...,\ \Tilde{p}_m^\prime$. 
We follow InfoNCE and use $\mathcal{L}_{CO}$ 
for training:

\begin{equation}
\label{eq:l_co}
   \mathcal{L}_{CO} = -log(\frac{exp(f_p\cdot f_{\Tilde{p}^\prime}/\tau)}{\sum_{i=1}^{m}exp(f_p\cdot f_{\Tilde{p}_i^\prime}/\tau)})
\end{equation}

In the meanwhile, we empower the cross-object correspondence with agnosticism to deformations.
We sample the observation $O^\prime$ of the first garment $O$ in another deformation state as described in Section~\ref{sec:method-cd}, and train its Cross-Deformation Correspondence using $\mathcal{L}_{CD}$ in Equation~\ref{eq:l_cd},
together with $\mathcal{L}_{CO}$ in Equation~\ref{eq:l_co} to make the learned representations aware of both cross-deformation and cross-object point-level correspondence.

\subsection{Coarse-to-fine Correspondence Refinement}
\label{sec:method-c2f}

Although above framework can learn the general distributions of all points' representations using offline randomly collected data, 
some difficult details (such as the boundaries between the folded sleeve on the garment body) should be paid more attention by the model, and there may exist inaccurate representations on some points or areas. The above phenomenon is also demonstrated in previous dense correspondence learning studies for 3D objects~\cite{hu2021efficient, halimi2019unsupervised, vestner2017efficient}.

Therefore, we propose the Coarse-to-fine (C2F) Correspondence Refinement procedure to make the model more focused on difficult points on the garment, and eliminate inaccurate predictions,
by refining the offline trained model using its online prediction failures.

Specifically,
for a certain garment, we sample a point $p$ on the observation $O$ in one deformation state, predict its point-level correspondence score on $O^\prime$ in another deformation state, with $p^\prime$ as corresponding point of $p$.
We collect points $\{p_1^\prime, p_2^\prime, ..., p_r^\prime\}$ that meet the following requirements: 
\begin{itemize}
    \item 
their correspondence scores are higher than a correspondence thresh $\alpha$; 
    \item 
their distances to $p^\prime$ in the canonical (flat) state (denoted as $d_{p_i^\prime}$ for $p_i^\prime$) are longer than a distance thresh $\beta$.
\end{itemize}
These points are prediction failures of the model trained on offline data.
We augment InfoNCE by distance (failure points farther away from $p^\prime$ will receive more penalties to take more focus) to refine the model on them:

\begin{equation}
\label{eq:l_c2f}
   \mathcal{L}_{C2F} = -log(\frac{exp(f_p\cdot f_{p^\prime}/\tau)}{\sum_{i=1}^{r} d_{p_i^\prime} \cdot exp(f_p\cdot f_{p_i^\prime}/\tau)})
\end{equation}
To prevent the model from forgetting the knowledge in offline data,
in this procedure,
we simultaneously train the model using offline data ($\mathcal{L}_{CO}$ and $\mathcal{L}_{CD}$) and online predictions ($\mathcal{L}_{C2F}$).

\subsection{From \textit{Topological} to \textit{Functional}: Few-shot Adaptation for Downstream Tasks}
\label{sec:method-functional}

While such topological structure of the above learned correspondence is significantly aligned with cross-object manipulation policy,
the point functionalities in different tasks may differ to some extent,
and thus could not be adequately reflected by a fixed representation.
To adapt the learned topological correspondence to be functional for different downstream tasks,
we propose the few-shot adaptation.

With the trained model, for a certain task and a functional action point (such as the pick point on the left sleeve for folding)
we annotate $l$ ($l \leq 5$) points $p_1, p_2, ..., p_l$ on $l$ observations $O_1, O_2, ..., O_l$ of different garments in different deformations, and fine-tune the trained model to make $f_{p_1}, f_{p_2}, ..., f_{p_l}$ to be the same using InfoNCE loss similar to Equation~\ref{eq:l_co} (replacing the \textit{topological} correspondence point with the \textit{functional} correspondence point as the positive sample).
Consequently,
the model can generally keep the topological information while become more aware of the functional information of the certain downstream task.

\subsection{Manipulation Policy Generation}
\label{sec:method-policy}

As shown in Figure~\ref{fig:teaser} and described in the last paragraph of Section~\ref{sec:pro}, for novel garments over different downstream tasks,
we can easily generate manipulation policies by selecting the picking and placing points that are most close to the demonstrations in the correspondence space.
More details of policy generation for diverse representative downstream tasks are described in Section~\ref{sec:policy}.

\subsection{Network Architectures and Training Strategy}
\label{sec:network}

Segmentation-version PointNet++~\cite{qi2017pointnet++} is used as the backbone feature extractor $\mathbf{F}$ that takes the point cloud observation $O$ as input to extract per-point features.
The per-point features are directly used to calculate correspondences.

We set batch size to be 32.
In each batch,
we sample 32 garment pairs. For each garment pair, we sample 20 positive positive point pairs, and 150 negative point pairs for each positive point pair.
Therefore,
in each batch, $32 \times 32 \times 20$ data will be used to update the model.
During the Correspondence training stage,
we train the model for 40,000 batches.
During Coarse-to-fine Refinement,
we train the model for 100 batches.
During Few-shot Adaptation,
we slightly refine the model using 5 demonstration data.

\section{Experiment}

\subsection{Simulation and Dataset}

We build our simulation environment based on the PyFleX bindings~\cite{li2018learning, lin2021softgym, xu2022dextairity}
to Nvidia FleX~\cite{macklin2014unified},
equipped with 3 kinds of garments, covering 500 tops (including shirts, hoodies, jumpers and etc.), 600 trousers and 600 dresses with diverse shapes, from the large-scale CLOTH3D dataset~\cite{bertiche2020cloth3d}.
An extra rack is loaded.

\subsection{Tasks and Metrics}

We evaluate our method over 3 different representative garment manipulation tasks:

\begin{itemize}
\item \textbf{Unfolding} that unfolds garments at random deformations to be flat. The unfolding succeeds when the coverage area of the unfolded garment exceeds a bar~\cite{ha2022flingbot}.
The two sub-tasks, \textbf{Unfold-RAND} and \textbf{Unfold-DROP},
respectively denote the garment initial states are generated by a few random actions or by dropping (more realistic).

\item \textbf{Folding} that folds garments. A folding succeeds when the Intersection-over-Union (IOU) between the target and the folded garments exceeds a bar~\cite{canberk2022clothfunnels, xue2023unifolding}.
\textbf{Fold-FLAT} and \textbf{Fold-FLING}
respectively denote garment initial states are perfectly flat or generated by flinging (more realistic).

\item \textbf{Hanging} that hangs garments on the rack, with the successful rate metric~\cite{chen2023learning}.
\textbf{Hang-RAND} and \textbf{Hang-FLING}
respectively denote garment initial states are generated by a few random actions or by flinging (more realistic).

\end{itemize}

\subsection{Baselines}

For \textbf{Folding}, we compare with ClothFunnels~\cite{canberk2022clothfunnels} that learns keypoints (\emph{e.g.}, endpoints of two sleeves, endpoints of the garment bottom line) from large-scale human-annotated data, pick and place keypoints step by step to fold garments.
 
For \textbf{Hanging}, we compare with GCSR~\cite{chen2023learning} that detects Structural Regions for manipulation (collars for hanging). 
Besides, we compare with DefoAfford~\cite{wu2023learning} that learns point-level actionable affordance scores for accomplishing the task and selects the best point for interaction.

For \textbf{Unfolding}, we compare with FlingBot~\cite{ha2022flingbot} that predicts the garment coverage area after flinging each two-point grasp pair, and selects the best pair for fling.
Besides, we compare with DefoAfford as it demonstrates the capability to unfold fabrics using one gripper.

\subsection{Manipulation Policy Generation}
~\label{sec:policy}
\vspace{-5mm}

For \textbf{Unfolding}, we use fling~\cite{ha2022flingbot} as the action for its quickness in unfolding garments using only a few steps.
It picks up 2 points simultaneously with 2 arms to lift the garment, pull the 2 points apart to stretch the garment, fling the garment and place it on the workplace.
As some keypoints for flinging may be occluded, we design 4 candidate pick pairs (\emph{e.g.}, the 2 endpoints of the shoulder, the 2 endpoints of the bottom line),
and select pick point pair of the observation with the highest correspondence to designed candidates.
We execute the fling action for at most 3 steps.

For \textbf{Folding}, we pick and place keypoints (defined in ClothFunnels~\cite{canberk2022clothfunnels}) on the garments to step by step fold garments with one or two arms.
Given the pick and place points sequence in the demonstration,
we select their closest pick and place points on each unseen object in the correspondence space and execute the pick-place action sequence.
Moreover,
to demonstrate the dense representations can facilitate multiple manipulation strategies with slight human annotations,
we show 4 folding strategies achieved by our method using one or few-shot demonstrations (Figure~\ref{fig:folding}).

For \textbf{Hanging}, we pick the point $p_{pick}$ that is most close to the demonstration pick point $\hat{p}_{pick}$, pull the garment up, and place it on the rack.

\subsection{Results and Analysis}

\begin{figure}[t]
  \centering
   \includegraphics[width=1\linewidth]{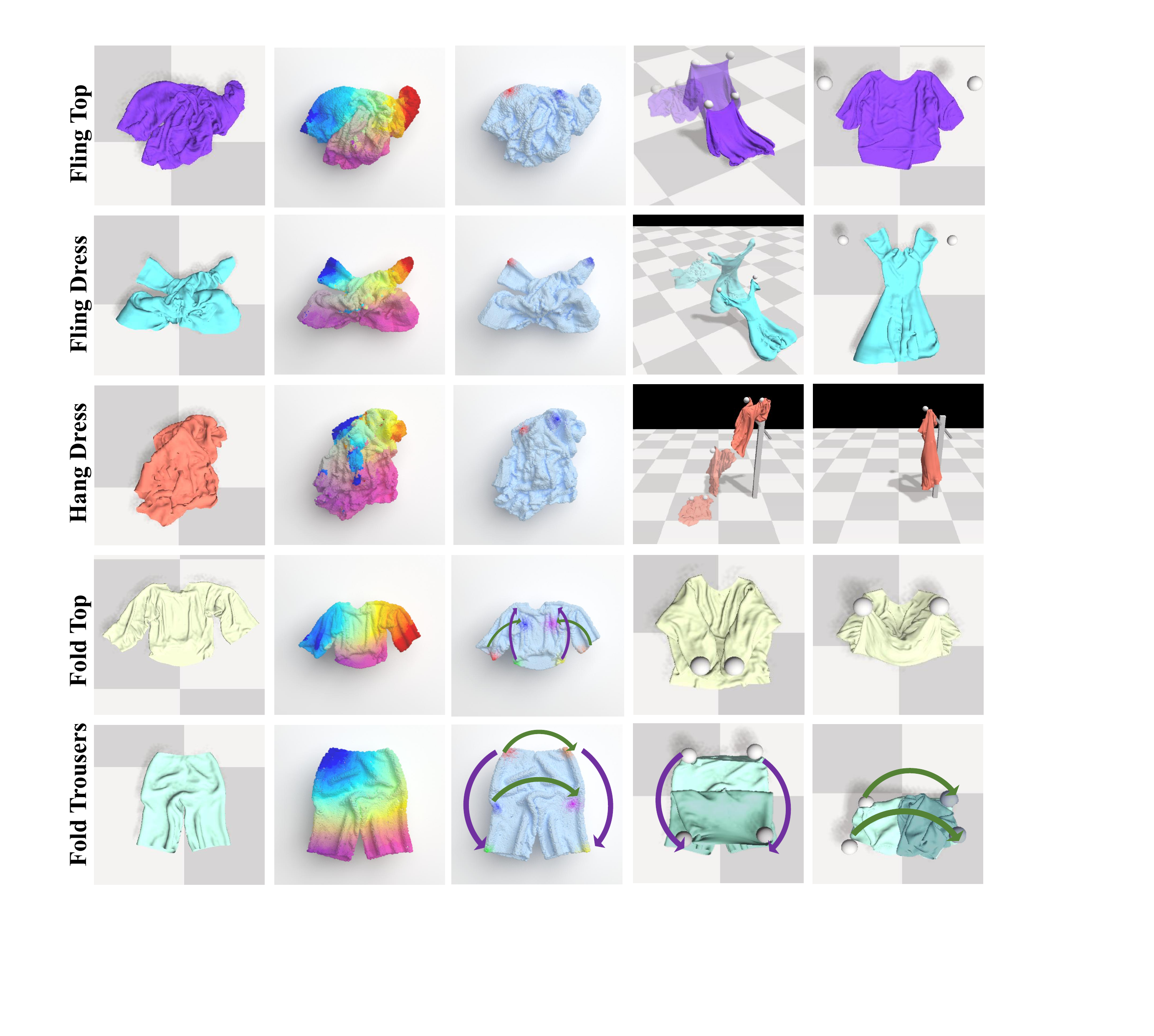}
   \caption{\textbf{ Correspondence Guided Manipulation on Different Garment Types and Tasks.} 
   From left to right:
   observation, correspondence, manipulation points (colored points) selected using correspondence to demonstrations and the
   manipulation action.
   }
   \label{fig:sequence}
\end{figure}

\begin{figure}[t]
  \centering
   \includegraphics[width=1\linewidth]{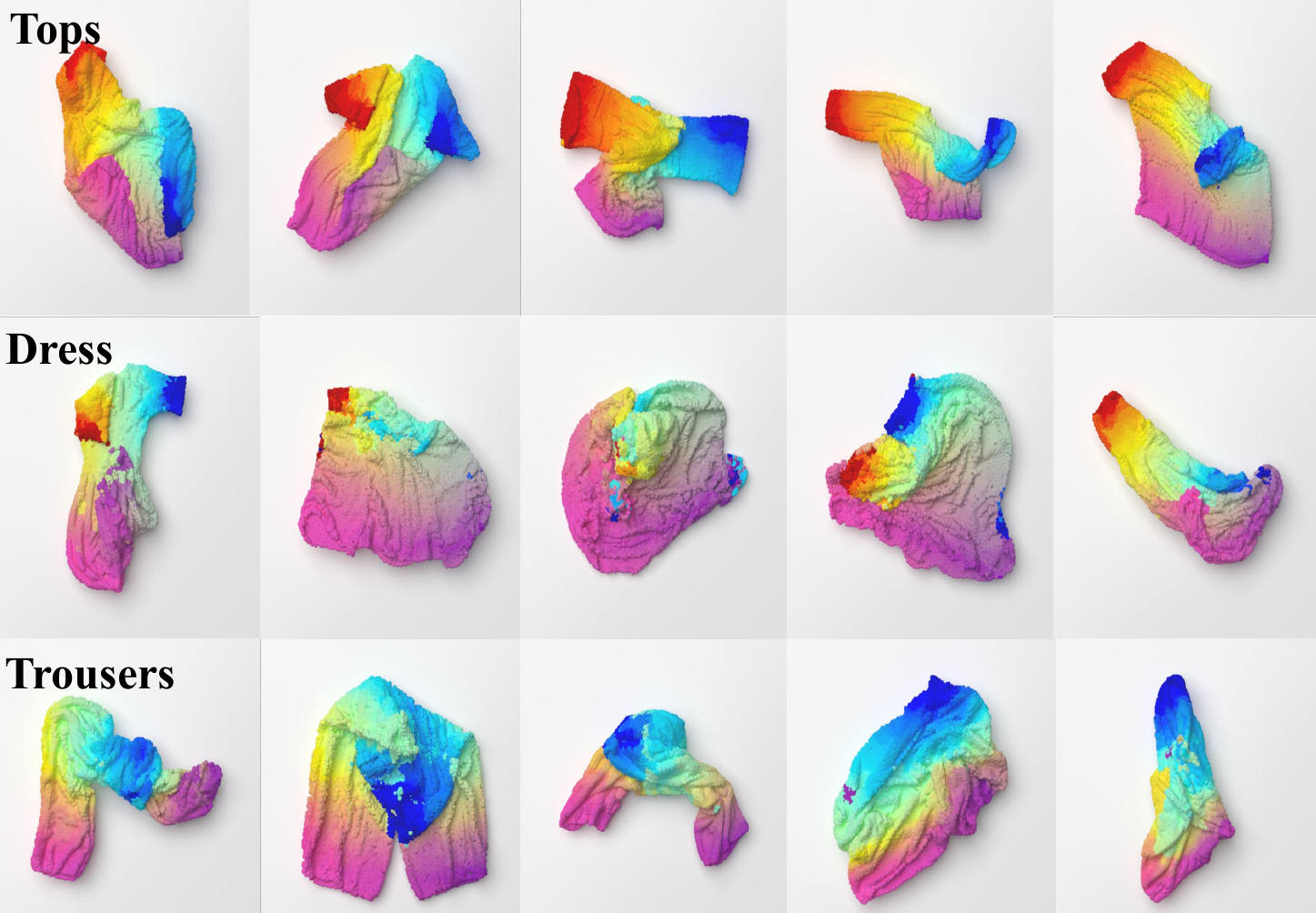}
   \caption{\textbf{Learned Dense Visual Correspondence.} For each category, we show correspondence for 5 objects in different deformations. Color similarity denotes correspondence similarity.}
   \label{fig:correspondence}
\end{figure}

Table~\ref{tab:unfold},~\ref{tab:fold} and~\ref{tab:hang} present quantitative comparisons with baselines. 
Figure~\ref{fig:correspondence} shows the learned correspondence on different garments shapes and deformations.
Figure ~\ref{fig:sequence} demonstrates the manipulation actions guided by correspondence.

\begin{table}
\begin{center}
    \begin{tabular}{lcccc}
     \toprule
    Method & Unfold-RAND (\%) & Unfold-DROP (\%) \\ \midrule \midrule
    FlingBot                & 80.3 / 82.1 / 79.9    & 84.3 / 86.7 / 81.9 
    \\  \midrule
    DefoAfford         & 70.1 / 63.9 / 61.7    & 73.8 / 60.9 / 66.7 
    \\  \midrule
    Ours        & \textbf{83.6} / \textbf{86.9} / \textbf{81.6}  &        \textbf{85.3} / \textbf{88.1} / \textbf{83.6}    \\  \bottomrule
    \end{tabular}
  \caption{\textbf{Results for Unfolding.} Numbers in the first / middle / last denote results for top / dress / trouser (the same below).}
  \label{tab:unfold}
\end{center}
\end{table}

\begin{table}
    \begin{tabular}{lcccc}
     \toprule
    Method & Fold-FLAT (\%) & Fold-FLING (\%) \\ \midrule \midrule
    ClothFunnels        & 82.2 / 83.9 / 79.6    & 61.7 / 63.5 / 60.3 \\ \midrule 
    UniFolding    & \textbf{83.5} / 82.9 / 81.6    & \textbf{78.7} / 81.5 / 78.6
    \\  \midrule
    Ours   & \textbf{83.5} / \textbf{84.0} / \textbf{83.3}   &    77.9 / \textbf{82.5} / \textbf{81.3}   \\  \bottomrule
    \end{tabular}
  \caption{\textbf{Results for Folding.}}
  \label{tab:fold}
\end{table}
    
\begin{table}
\begin{center}
    \begin{tabular}{lcccc}
     \toprule
    Method & Hang-RAND (\%) & Hang-FLING (\%) \\ \midrule \midrule
   GCSR      & 78.5 / 72.3 / 77.9   & 81.2 / 80.9 / 78.7
    \\  \midrule
    DefoAfford         & 73.4 / 69.2 / 71.6   & 79.4 / 76.4 / 73.8  
    \\  \midrule
    Ours        & \textbf{81.9} / \textbf{77.4} / \textbf{83.3}  &        \textbf{83.8} / \textbf{89.6} / \textbf{81.5}    \\  \bottomrule
    \end{tabular}
  \caption{\textbf{Results for Hanging.}}
  \vspace{-6mm}
  \label{tab:hang}
\end{center}
\end{table}

\begin{figure}[t]
  \centering
  \includegraphics[width=0.5\textwidth]{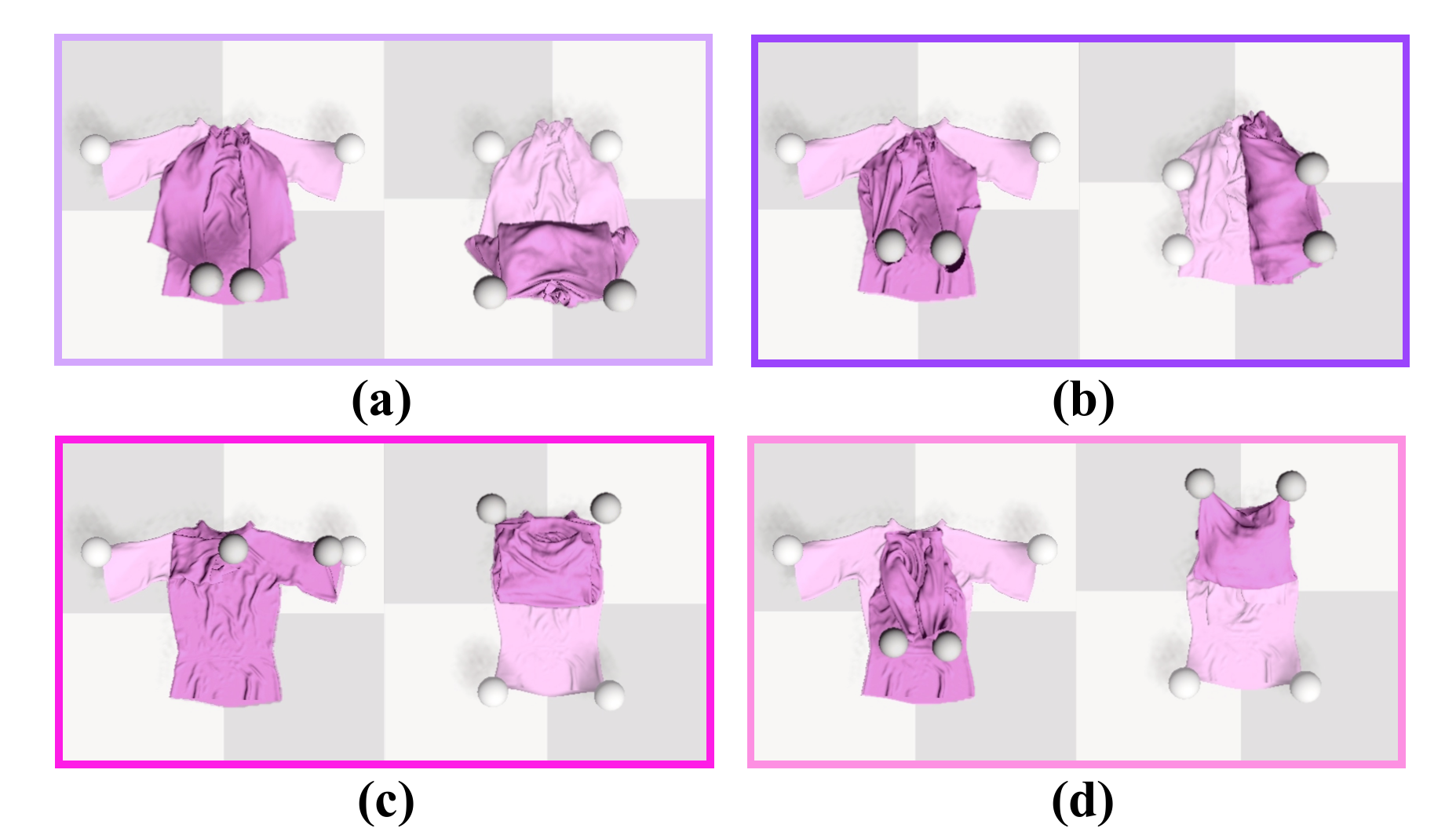}
  \vspace{-7mm}
   \caption{\textbf{Visualization of Different Folding Policies.}}
   \label{fig:folding}
  \vspace{-5mm}
\end{figure}

For folding,
as ~\textbf{ClothFunnel}'s keypoint detection model is trained on garments in fully unfolded states,
it is difficult to generalize to garments unfolded using FlingBot or our method,
as garments could not be perfectly unfolded into a fully flat state.
In contrast,
as our method is trained with the awareness of self-deformations,
it is easier to detect such keypoints in diverse garment states.
While ~\textbf{UniFolding} trains using and thus works well on deformed states,
it is designed for the specific folding task.
In contrast, our method can facilitate multiple downstream tasks.
Furthermore,
as shown in Figure~\ref{fig:folding}, our method can work well on different ,
while policies trained on large-scale annotated data cannot easily generalize to novel manipulation methods.

For unfolding,
DefoAfford cannot perform well in 3 steps as it only utilizes one gripper.
For FlingBot, although it is trained using large-scale different states and interactions,
without training on many garments, it cannot generalize well to novel garments.
Besides, it is costly in time and computing resources, in that it requires separately training 96 models to generate affordance maps in 12 garment rotation types and 8 garment scale types, and then selecting the best pick points pair in all the 96 rotation-scale combinations.
In contrast,
as the learned dense correspondence is aware of different garment scales and rotations,
we can directly use one model to select the grasp points for flinging.

For the comparison with GCSR in folding,
as it requires collar detection as Structural Regions,
it cannot perform well on garment states where collars are occluded.

It is worth mentioning that,
while the baselines are mostly designed for specific tasks and may pose requirements to garment initial states,
our proposed dense visual correspondence is an \textbf{unified} representation for garments, and thus makes it easy to generate policies for \textbf{multiple downstream tasks} on \textbf{different garment deformations}.

\subsection{Ablation Studies}

We compare our method with the ablated versions on the representative folding task, to demonstrate the effectiveness of our method's different components: 

\begin{itemize}
\item 
\textbf{Ours w/o CO}: our method without learning the cross-object correspondence. 
This version is similar to ~\cite{ganapathi2021learning}
that learns correspondence and policies on square-shaped fabrics and similar T-shirts with very short sleeves.

\item 
\textbf{Ours w/o CD}: our method without learning the cross-deformation correspondence.

\item 
\textbf{Ours w/o C2F}: our method without C2F Refinement.

\item 
\textbf{Ours w/o FA}: our method without few-shot adaptation on different downstream tasks.
\end{itemize}

Table~\ref{tab:ablation} shows quantitative comparisons with ablations. Clearly each component improves our method's capability.

\begin{figure}[t]
  \centering
  \includegraphics[width=0.5\textwidth]{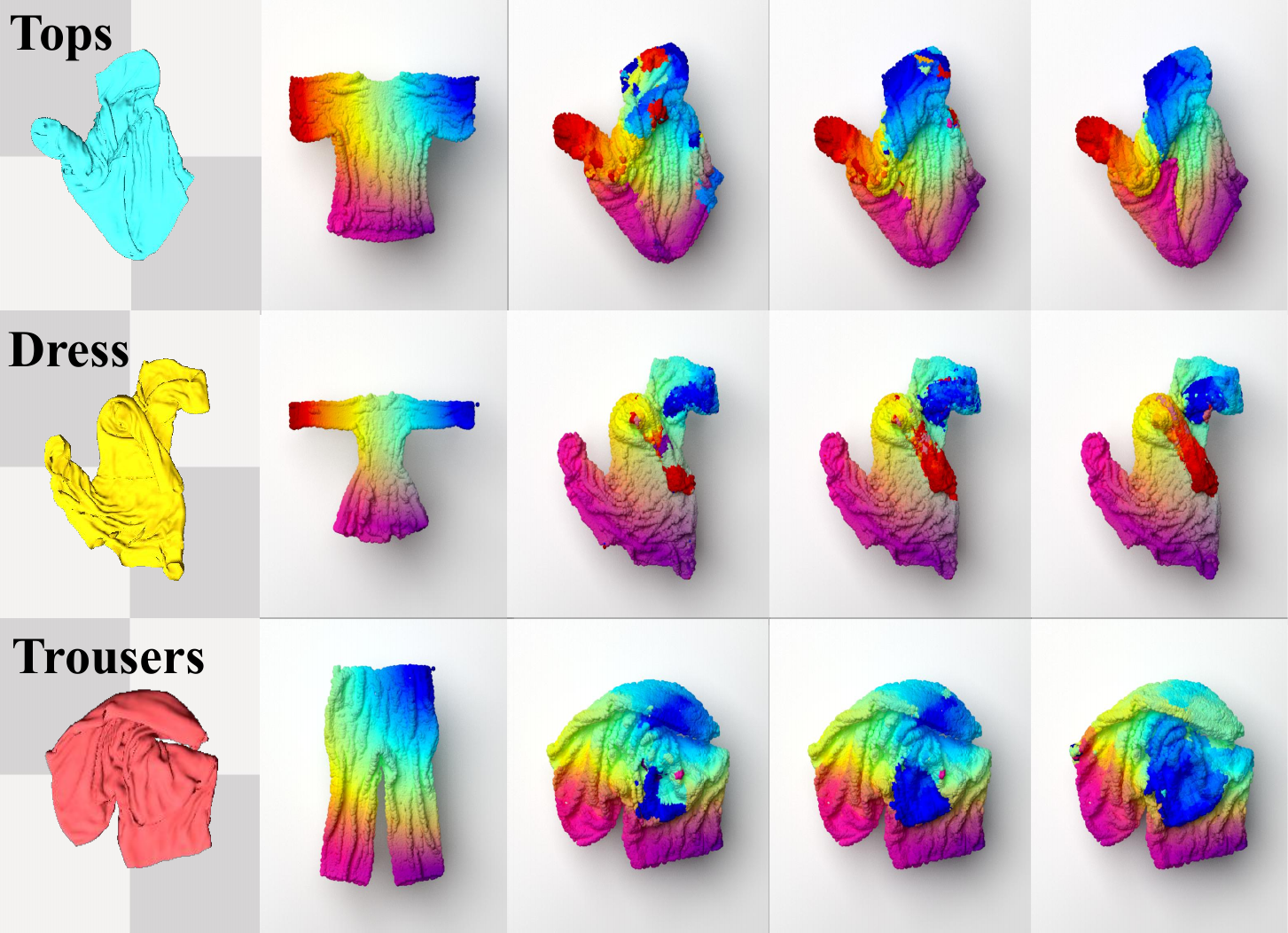}
   \caption{\textbf{Coarse-to-fine Refinement}. From left to right: the garment, correspondence in its flat pose, initial correspondence, correspondence after 50 and 100 refinement batches.}
   \label{fig:c2f}
\end{figure}

As shown in Figure~\ref{fig:c2f}, Coarse-to-fine Refinement eliminates many inaccurate areas of correspondence prediction, and makes the boundaries of garment parts more clear.

\begin{figure}[t]
  \centering
  \includegraphics[width=0.5\textwidth]{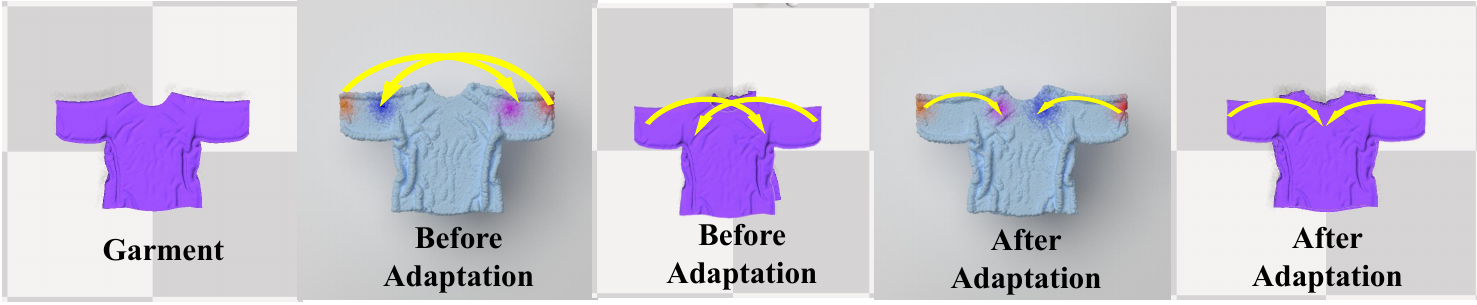}
   \caption{\textbf{Ablation on Few-shot Adaptation.}}
   \label{fig:fa}
  \vspace{-5mm}
\end{figure}

As shown in Figure~\ref{fig:fa},
after few-shot adaptation,
the manipulation points tend to be more functional,
and thus the folded garments become more organized.

\begin{table}
\begin{center}
    \begin{tabular}{lcccc}
     \toprule
    Method & Top (\%) & Dress (\%)  & Trouser (\%) \\ \midrule \midrule
   Ours w/o CO        & 75.3   & 70.8 & 77.6 
    \\  \midrule
    Ours w/o CD         & 63.9   & 65.2 & 59.4 
    \\  \midrule
   Ours w/o C2F                 & 78.0   & 80.3 & 76.7 
    \\  \midrule
    Ours w/o FA         & 81.9   & 82.3 & 78.6 
    \\  \midrule
    Ours        & \textbf{83.6}  &        \textbf{86.9} & \textbf{81.6}    \\  \bottomrule
    \end{tabular}
  \caption{\textbf{Ablation Studies.}}
  \vspace{-5mm}
  \label{tab:ablation}
\end{center}
\end{table}

\subsection{Real-world Evaluation}

\begin{figure}[t]
  \centering
   \includegraphics[width=1\linewidth]{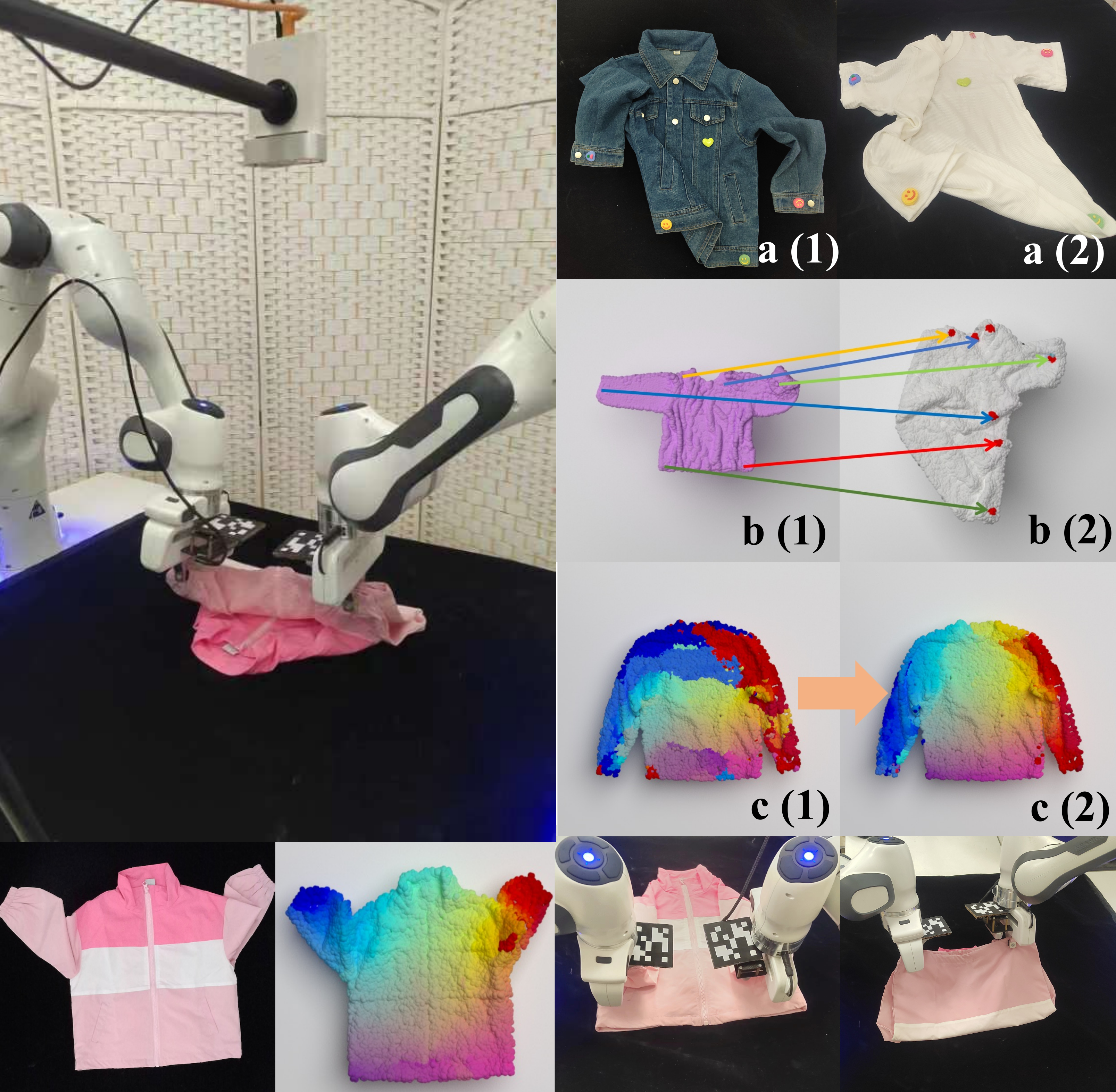}
   \caption{\textbf{Real-world Setup and Experiments.}}
   \label{fig:real}
\end{figure}

\textbf{Setup.} As shown in Figure~\ref{fig:real}, our real-world experiment setup consists of two Franka Panda robot arms, and a Microsoft Azure Kinect camera (which has demonstrated high-precision with slight noises for robotic manipulation~\cite{cheng2023learning, ning2023where2explore}) capturing top-down point cloud.
We use Segment Anything (SAM)~\cite{kirillov2023segment} to segment the garment from the scene and project the segmented image with depth to point cloud. 

Please refer to the supplementary materials for more details and videos of real-world manipulations.

To align the scanned point cloud with those in simulation,
we annotate a few skeleton points on 3 real-world garments (Figure~\ref{fig:real}, a), 
 with the correspondence between annotated skeleton points and those in simulation shown in b(1) and b(2),
and fine-tune the pre-trained model by pushing close the representations of skeleton points on garments in simulation and the real world using InfoNCE. 
The correspondence is adapted from c(1) to c(2) in Figure~\ref{fig:real}.

We use 5 real world different-shaped tops, each conducting folding on 3 different initial deformations, and report the number of successful executions in Table~\ref{tab:real}.

\begin{table}
\begin{center}
    \begin{tabular}{lcccc}
     \toprule
    Method & ClothFunnel & UniFolding & Ours \\ \midrule   \midrule
    Fold        &  8 / 15  &        10 / 15 &  \textbf{11} / 15  \\  \bottomrule
    \end{tabular}
  \caption{\textbf{Real-world Evaluation on Folding.} }
  \label{tab:real}
  \vspace{-7mm}
\end{center}
\end{table}

\section{Conclusion}

We propose to learn dense visual correspondence for diverse garment manipulation tasks with category-level generalization using only a few annotations.
We first train topological correspondence self-supervisedly using self-play and garment skeletons,
and then fine-tune it using few-shot demonstrations to transform the topological correspondence to be functional to different downstream tasks.
Extensive experiments demonstrate the superiority of our method.

\vspace{-3mm}
\section{Acknowledgment}
\vspace{-3mm}

This work was supported by National Natural Science Foundation of China (No. 62136001) and National Youth Talent Support Program (8200800081).
We thank Yangyi Ye, Kunqi Xu and Haotong Zhang for assisting real-world experiments, Yang Tian and Jiyao Zhang for discussions.

\appendix

\section{More Details of Real World Setup}

\subsection{Setup}

We use two Franka Emika Panda~\cite{franka-emika-panda} robot arms to execute the manipulation in the real world. To control the robot, we use the LibFranka library~\cite{libfranka}. 
We mount a Microsoft Azure Kinect camera (which has demonstrated high-precision with slight noises for robotic manipulation~\cite{cheng2023learning, ning2023where2explore}) capturing top-down RGBD image.

\subsection{Garment Segmentation}

\begin{figure}[t]
  \centering
   \includegraphics[width=1\linewidth]{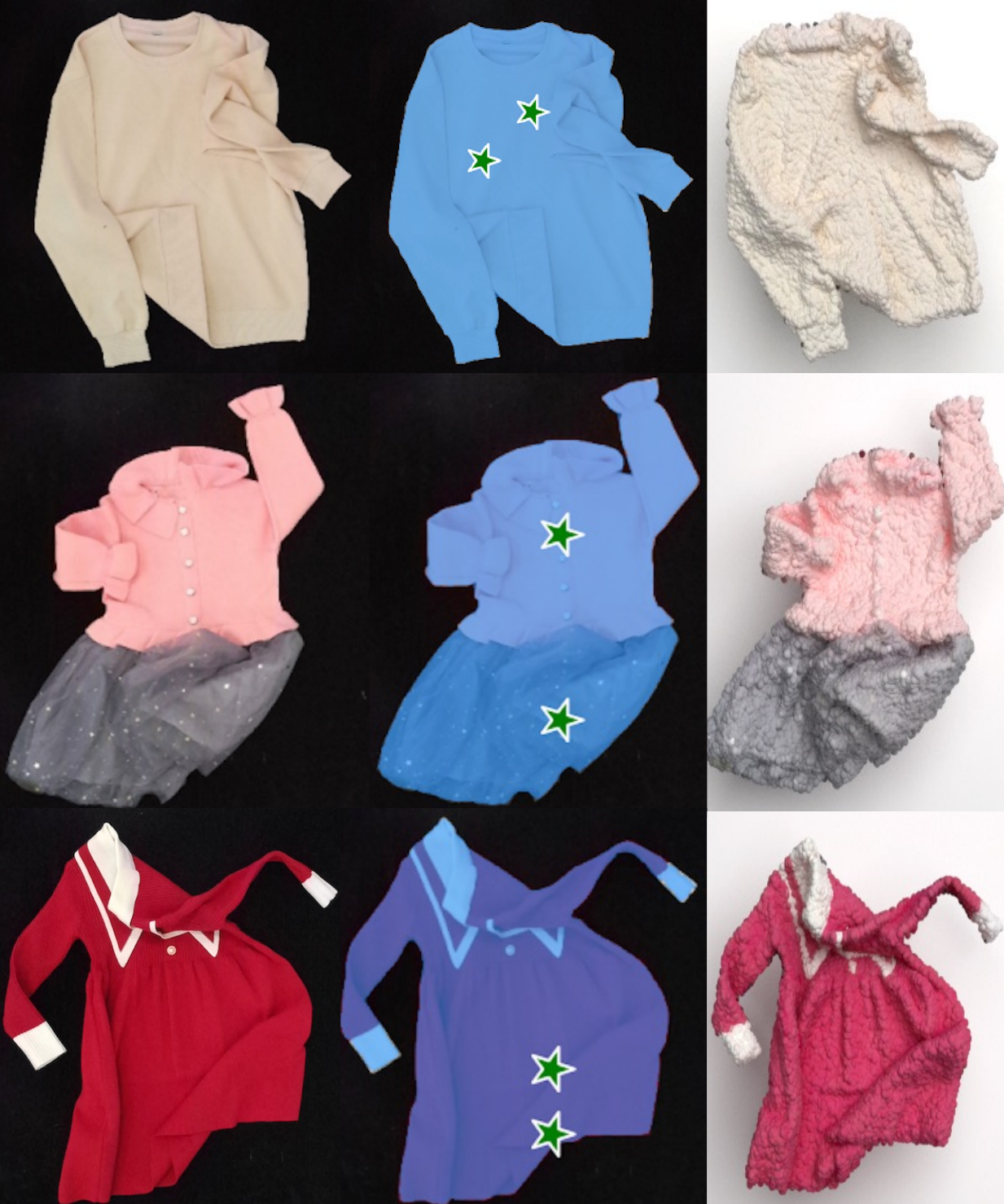}
   \caption{\textbf{RGB observation, SAM and projected Point Cloud.}
   }
   \label{fig:sam}
\end{figure}

As shown in Figure~\ref{fig:sam}, we use the default configuration of Segment Anything (SAM)~\cite{kirillov2023segment} to segment the garment from the scene in the RGB image,
and project the corresponding depth image into the point cloud.

\subsection{Point Cloud Alignment}

\begin{figure}[t]
  \centering
   \includegraphics[width=1\linewidth]{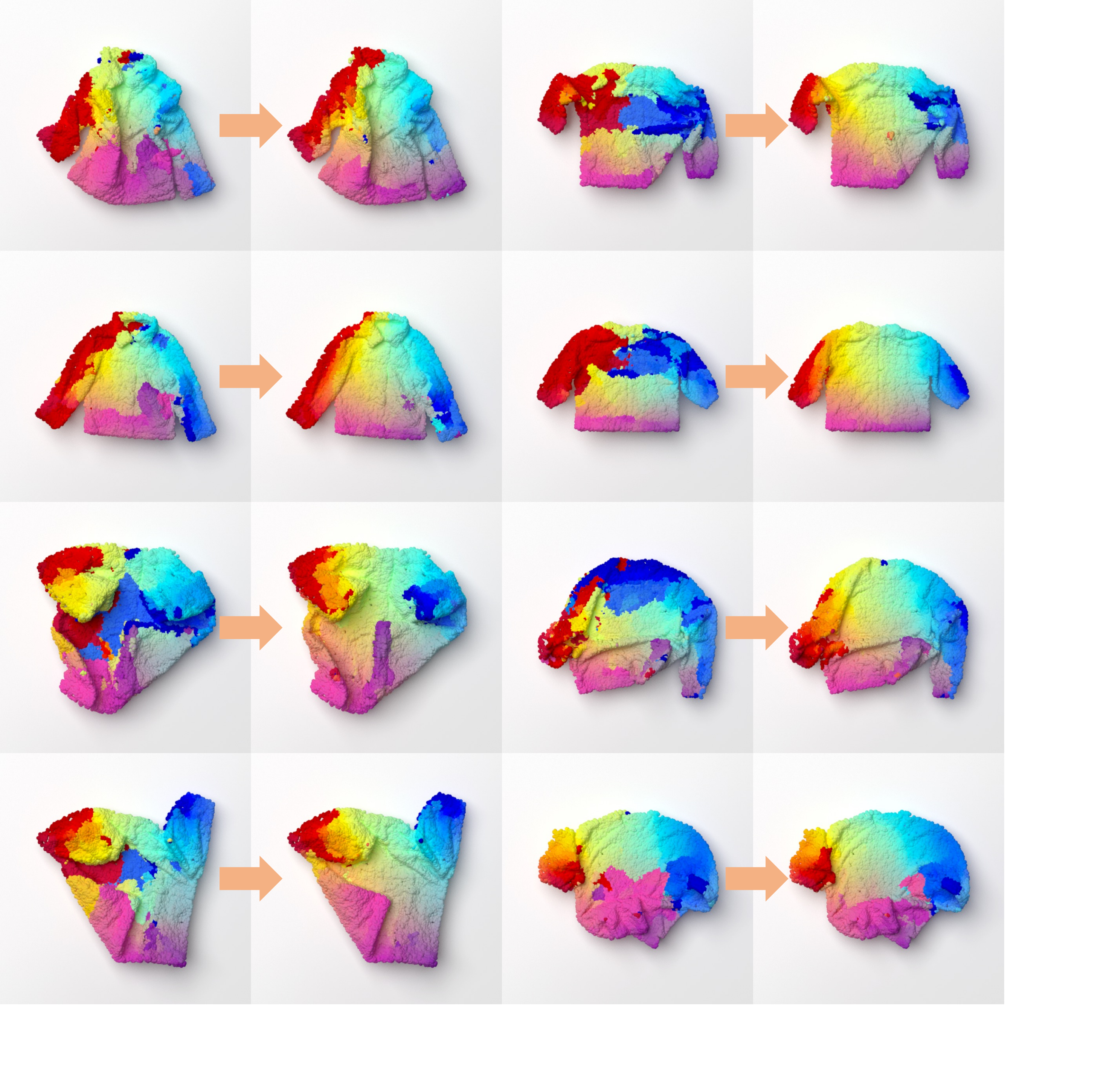}
   \caption{\textbf{More Results on Real-world Adaptation.}
   }
   \label{fig:tune}
\end{figure}

To align the scanned point cloud with those in simulation,
we annotate a few skeleton points on 3 real-world garments, 
 with the correspondence between annotated skeleton points,
and fine-tune the pre-trained model by pushing close the representations of skeleton points on garments in simulation and the real world using InfoNCE. 
Figure~\ref{fig:tune} shows more results on the Correspondence after the Point Cloud Alignment, apart from the figure in the main paper.

\begin{figure}[t]
  \centering
   \includegraphics[width=1\linewidth]{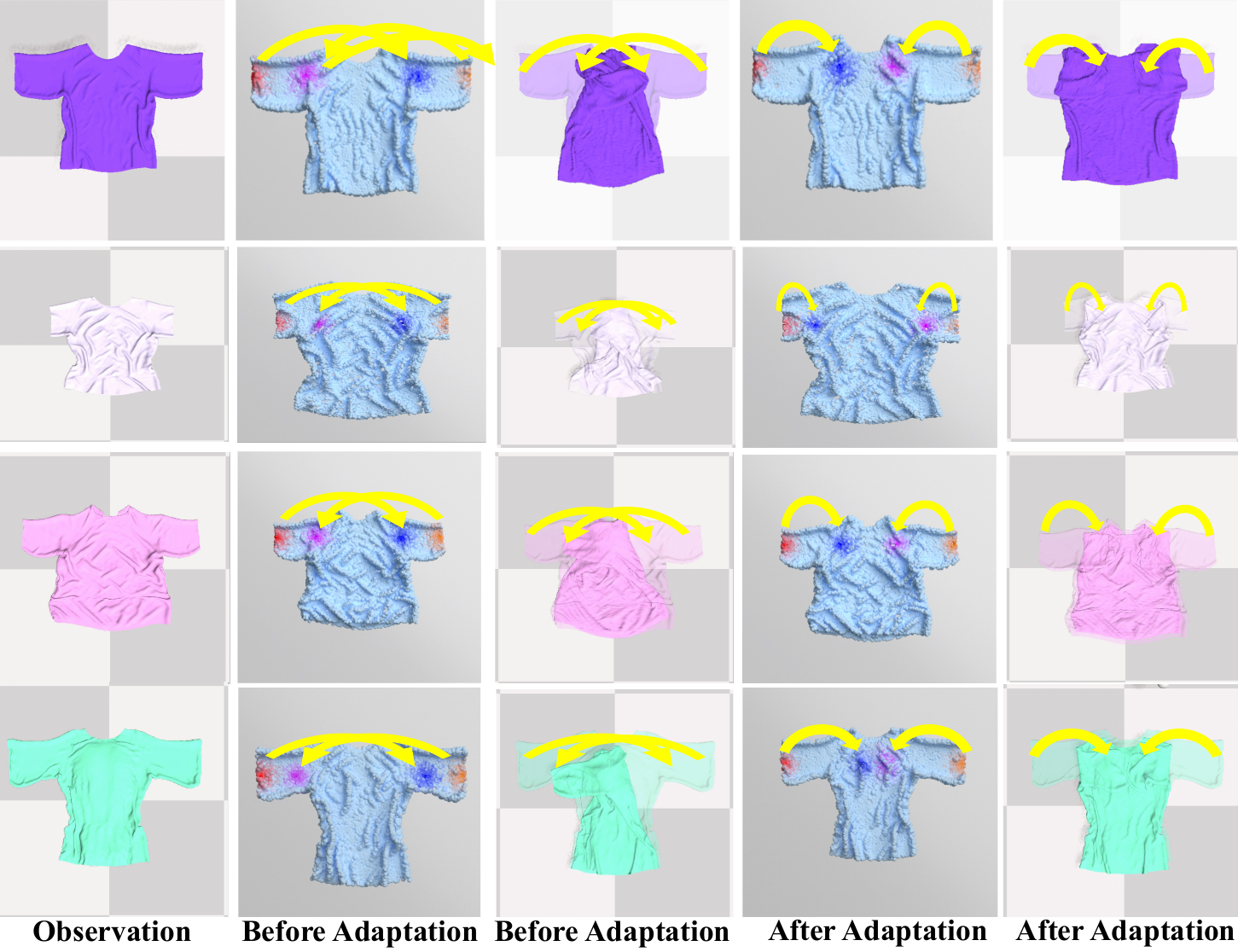}
   \caption{\textbf{More Results of Few-shot Adaptation.}
   }
   \label{fig:supp_fewshot}
\end{figure}

\section{Hyper-parameters Selection}

We set batch size to be 32.
In each batch,
we sample 32 garment pairs. For each garment pair, we sample 20 positive point pairs, and 150 negative point pairs for each positive point pair.
Therefore,
in each batch, $32 \times 32 \times 20$ data will be used to update the model.
During the Correspondence training stage,
we train the model for 40,000 batches.
During Coarse-to-fine Refinement,
we train the model for 100 batches.
During Few-shot Adaptation,
we slightly refine the model using 5 demonstration data.
Besides,
we set the number of skeleton pairs to be 50.

\begin{figure*}[t]
  \centering
   \includegraphics[width=1\linewidth]{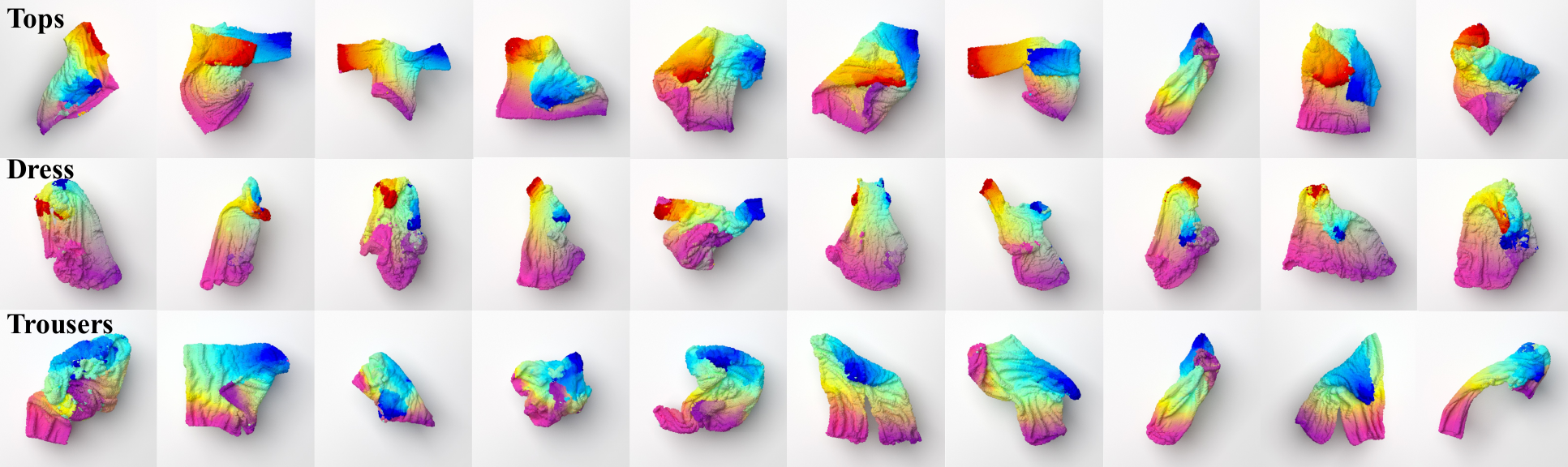}
   \caption{\textbf{More Results of Dense Correspondence.}}
   \label{fig:supp_corr}
\end{figure*}

\section{Computational Resources}

We use PyTorch as our Deep Learning framework. 
Each experiment is conducted on an RTX 3090 GPU, and consumes about 22 GB GPU Memory for training. It takes about 12 hours to train the Coarse Stage, with 1-2 hours of Coarse-to-fine Refinement and 0.5 hour's Few-shot Adaptation.

\section{More Results of Dense Correspondence}
Figure~\ref{fig:supp_corr} shows more results of dense correspondence.

\section{More Results of Few-shot Adaptation}
Figure~\ref{fig:supp_fewshot} shows more results of few-shot adaptation.

{
    \small
    \bibliographystyle{ieeenat_fullname}
    \bibliography{main}
}


\end{document}